# COMBINING PARAMETRIC LAND SURFACE MODELS WITH MACHINE LEARNING

[1,2,4]Craig Pelissier, [3]Jonathan Frame, [5,2,3]Grey Nearing

[1]NASA Goddard Space Flight Center, Greenbelt, MD, [2]University of Maryland Baltimore County, Baltimore, MD, [3]University of Alabama, Tuscaloosa, AL, [4]Science Systems Applications Inc., Lanham, MD, [5]Upstream Tech, Public Benefit Corporation, Alameda CA

## ABSTRACT

A hybrid machine learning and process-based-modeling (PBM) approach is proposed and evaluated at a handful of AmeriFlux sites to simulate the top-layer soil moisture state. The Hybrid-PBM (HPBM) employed here uses the Noah land-surface model integrated with Gaussian Processes. It is designed to correct the model only in climatological situations similar to the training data else it reverts to the PBM. In this way, our approach avoids bad predictions in scenarios where similar training data is not available and incorporates our physical understanding of the system. Here we assume an autoregressive model and obtain out-of-sample results with upwards of a 3-fold reduction in the RMSE using a one-year leave-one-out cross-validation at each of the selected sites. A path is outlined for using hybrid modeling to build global land-surface models with the potential to significantly outperform the current state-of-the-art.

## 1. INTRODUCTION

The value of Machine Learning (ML) in Earth Sciences is well established, and it is an active area of research across a wide variety of application domains. In hydrological modeling, we have already seen many instances where ML has *substantially* outperformed traditional process-based modeling (PBM) [e.g., 6-7, 9]. These results motivate the need for answers to important questions about the use of ML "black-box" models in science:

1. *How can we trust ML models?*
2. *How do we gain an understanding of the processes that drive the systems our ML models predict?*

The first question arises because ML employs flexible non-parametric models whose structure is constructed entirely from data. Traditional PBMs are constructed from some underlying physical principles that we *believe* are generally valid in any climatology. ML or data-driven models, on the other hand, can only be expected to do well in situations similar to the data the model was trained on. It is possible, and we hope, that ML models learn the fundamental underlying processes, but at this point we do not know of a way to satisfactorily determine if this is true. An obvious approach to circumvent this problem is to somehow combine our physical understanding with ML, and in fact this is an area actively being pursued [5, 10]. There are two approaches that might be employed: (1) imposing physical constraints on our ML models either analytically [e.g., 1] or through penalties imposed on the cost function, and (2) integrating PBMs with ML [e.g., 8]. In this work, we propose a method using the latter approach.

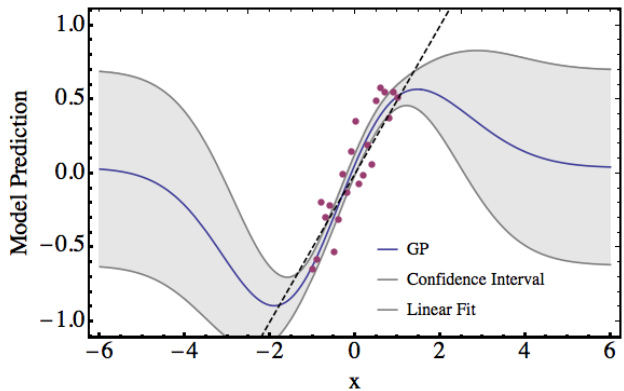

**Figure 1** Example of Gaussian Process (GP) regression (blue) on pseudo-noisy-linear data (purple) compared to a standard linear regression (black). Here we see that the zero mean GP prior causes the fit to extrapolate to zero outside of the data as opposed to parametric (linear) regression.

The second question is perhaps more difficult to answer, and to the authors knowledge, no satisfactory and universal solution exists at this time. However, before we can learn new physics from ML, we must first build reliable global ML models that captured the fundamental processes of the systems being modeled. Here we propose a potential approach, evaluate its performance modeling the soil moisture state, and present promising initial results.

## 2. COMBINING PROCESS-BASED MODELS WITH MACHINE LEARNING

To integrate PBMs with ML, we require the basic design principle that the ML model should only be active in situations that are climatologically similar to the training data, otherwise it should revert back to the PBM. For convenience, we will refer to such a model as a Hybrid-PBM (HPBM). The HPBM addresses (1) in Section 1, and in addition it addresses the important practical problem of having a "complete" training set. In some cases, Earth observational data may be extensive and in others less so. This may be true for several reasons such as limited sensor coverage, rare climatological events, or even unseen climatology induced by anthropogenic change. A HPBM allows you to make use of incomplete data sets by only incorporating observational data *when* it is available. When it is not available the model defaults back to the PBM which is out best estimate based on our physical understanding of the system.

To create a HPBM, we start by denoting our PBM by

$$x_{t+1} = f(x_t), \qquad (1)$$

where $x_t$ denotes the state-vector at time $t$, and we assume a discrete time index. The ML model is incorporated straightforwardly as

$$v_{t+1} = f(v_t) + ML(v_t) = \psi(v_t), \qquad (2)$$

where $v_t$ is used to distinguish the timeseries under the evolution of $\psi$ and $ML$ denotes some unspecified ML model. To adhere to our principle, we must have $ML(v_t) = 0$ whenever the state $v_t$ of the system is dissimilar to the available training data. This can be accomplished by placing a vanishing prior mean on the ML model in which case

$$E[\psi(v_t)] = f(v_t) + E[ML(v_t)] = f(v_t). \qquad (3)$$

This can be achieved naturally with Gaussian Processes (GPs), and is in fact, standard practice [11]. To highlight this, in Figure 1 we show the fit results on pseudo-noisy-linear data using GPs and a linear-fit model. Unlike the linear-fit model, GPs extrapolate to zero outside of the data. While this lack of extrapolability is often thought of as a weakness of GPs, here it is desirable behavior. For this reason, GPs are an ideal candidate to integrate ML an PBMs, and as such, is chosen as the ML model in this study. We then write our HPBM as

$$\psi(v_t) = f(v_t) + \mathcal{GP}(v_t). \qquad (4)$$

We note that although simple in form, the flexibility of GPs makes this model quite general and capable of capturing a large class of climatological responses.

### 3. GAUSSIAN PROCESS AUTOREGRESSION

If we replace the second term on the RHS of Eq. 4 with a random white noise, we have the equation of a classical data assimilation (DA) problem. In fact, the HPBM *is* a form of data assimilation; The ML model tries to correct the PBM using observational data. However, there is a key difference. In DA, we add random fluctuations to the model and use a Bayesian approach to correct the model based on observations. In this case, we add a *structured* corrective term and *learn* the model structural error though the error patterns between observations and model predictions. In this way, unlike classical DA, we create a *new* dynamics model that can assimilate observations both in- and out-of-sample. Just like classical DA, the non-linearity of the PBMs leads to intractable integrals that have to be evaluated either using Monte Carlo techniques or by making simplifying assumptions that make the integrals tractable. Since here we are trying to learn the model structural error, it is essential to retain the non-linearity of the GP, and linear approximations like the Kalman filter are not sufficient.

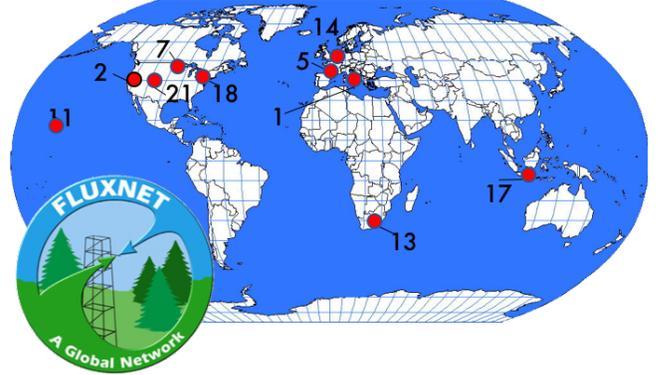

**Figure 2** Geographical locations of the ten AmerFlux sites used to test the results of the hybrid machine learning process-based model.

Our HPBM is a Gaussian Process State-Space-Model (GP-SSM), and there have been several recent advances related to training GP-SSMs. An approachable overview is given by Frigola [3]. Though ultimately GP-SSMs will have to be employed, to investigate the potential of this approach we circumvent the intractable integrals by ignoring the uncertainty propagation and performing autoregression. This approach is possible as long as one has observational pairs $(y_{t+1}, y_t)$ at the desired simulation timestep, and the quality of the data is sufficient to determine $\mathcal{GP}(v_t)$ with a small enough uncertainty. To meet these requirements, we test our approach at AmeriFlux tower sites where in-situ half-hourly data is available.

To train the autoregression model, we make the assumption that the input state $v_t = y_t$. In this case the input/output training pairs are $(y_{t+1} - f(y_t), y_t)$, see Equation 4. The "one-step-ahead" training set is generated as follows: (1) initialize $v_0 = y_0$, (2) set $v_t = y_t$, (3) compute $y_{t+1} - f(v_t)$ and record, and (4) repeat steps 2-3 for $i = 1, T$.

Often, as in the case of this work, one only observes part of the state-vector $v_t$. In this case, the unobserved part of the state vector is evolved according to the PBM *i.e.* the prediction from the PBM model for the unobserved part of the state-vector is used as input to the next timestep unaltered by the ML model. This process is then equilibrated by enforcing temporal periodic boundary conditions and running through the data several times. In this way, the unobserved portion of the state is equilibrated to the observed portion of the climatology.

Table 1 Performance results of the Hybrid-Process-Based-Model (HPBM) using the Noah hydrological model. The first column indicates the site locations, see Figure 2, and the second column shows the number of years of training data. The third and fourth columns show the cross-validated average performance of the Noah and HPBM, and the last column displays the percent improvement.

| Site | # Years Data | Noah <RMSE> | HPBM <RMSE> | % Improvement |
|---|---|---|---|---|
| 1 | 3 | 0.053 | 0.041 | 23 |
| 2 | 7 | 0.061 | 0.02 | 67 |
| 5 | 4 | 0.033 | 0.02 | 39 |
| 7 | 2 | 0.084 | 0.08 | 5 |
| 11 | 4 | 0.027 | 0.017 | 37 |
| 13 | 2 | 0.043 | 0.037 | 14 |
| 14 | 4 | 0.032 | 0012 | 62 |
| 17 | 4 | 0.04 | 0.036 | 10 |
| 18 | 2 | 0.04 | 0.013 | 68 |

## 4. RESULTS

### 4.1. Experimental setup

To test the HPBM, we used the Noah land surface model as our PBM and predicted the top-layer (5 cm) soil moisture state at a collection of AmeriFlux towers sites, see Figure 2. The training data consist of observed values of the top layer soil moisture, $y_t$, and forcing inputs, $u_t$, at half-hour intervals. The unobserved portion of the state-vector $s_t$, including lower level soil moisture states, is not corrected and is evolved solely by the PBM model. Denoting the top layer soil moisture by $\theta_t$, we write the state vector as $v_t = (s_t, u_t, \theta_t)$. The one-step-ahead training procedure in this case follows the prescription in Section 3 with step (1) replaced with $\theta_0 = y_0$ and step (2) with $v_t = (s_t, u_t, \theta_t)$.

To train the HPBM, we use the squared-exponential (SE) kernel [11]

$$K(x',x) = \sigma_f^2\, e^{-\frac{1}{2}(x'-x)^T \Theta^{-1}(x'-x)}, \quad (5)$$

and employ sparse-pseudo input GPs (SPGP) [12].

For each training set, we vary the number of pseudo inputs and perform several restarts to avoid bad local minima. To select the model with the best performance, it is important to measure the *dynamical* performance. To do this, we run the HPBM model on a test year, compute the RMSE, and choose the model with the lowest RMSE. A leave-one-out (LOO) cross-validation is performed over all available years at each site, and the results are reported in Section 5.

To carry out this and future work, we developed an efficient C++ High-Performance Computing (HPC) implementation of GPs capable of training on tens-of-millions of samples. This software package will be released under NASA's public license and made available to the scientific community throughout this work. All computer simulations and numerical work were carried out at the NASA Center for Climate Simulations.

### 5.2. Results

We identified nine AmeriFlux stations with a variety of different climates that have top layer soil moisture observations available. At each site, we first calibrate the Noah model on the training data using Shuffled Complex Evolution [2]. Then we create the one-step-ahead training sets and train the HPBMs. We did not use a separate test set since some of the sites did not contain enough data.

For each site, we select the model with the best performance on the validation year, cross-validate over all the years, and computed the average RMSE. The results are shown in Table 1. We see that all sites show an improvement with some sites achieving a 3-fold reduction in the RMSE. Site 7 saw no significant improvement, but it was only trained using one-year of data and is likely a result of those 2 years being climatologically different.

The HPBM models were trained by varying the number of pseudo-inputs in the SPGP with ten restarts for each pseudo-input. Altogether, for each training set, we produced ~100 sets of fit parameters. We found the regression performance was good and quite stable, but the *dynamical* performance in- and out-of-sample varied sometimes substantially. This is likely the result of ignoring the temporal uncertainty propagation, and we expect it will be resolved using GP-SSMs.

In Table 1, we show the results for the parameter sets that performed best out-of-sample. In practice, the model selection should not be based on the out-of-sample dynamical performance. While the in- and out-of-sample dynamical performance is well correlated, it was not true that the models that performed the best in-sample were the ones that performed best out-of-sample. One way to incorporate all the models to create a more stable predictor is to marginalize.

This can be accomplished by combining the GP posterior predictive distributions *via* model averaging [4],

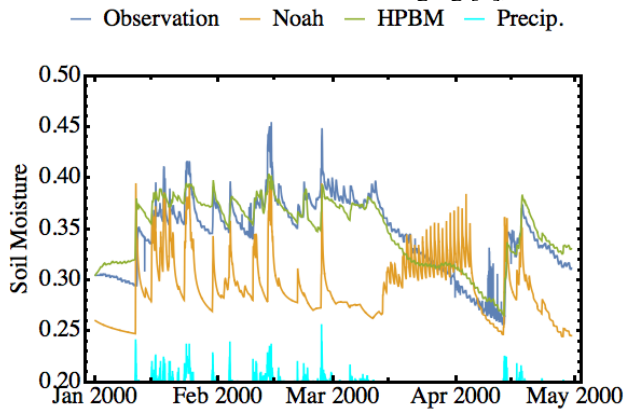

**Figure 3** A time series plot of the observations (blue), Noah (orange), HPBM (green), and relative precipitation (cyan). In the spring, the Noah model exhibits an oscillatory artifact that indicates a model structural error related to snowmelt. The HPBM is able to learn and correct this structural error which is easily seen by comparing to observations.

$$p(x^*|D) = \sum_i p(x^* \mid M_i, D)\, p(M_i|D), \quad (6)$$

which results in a posterior predictive mean,

$$\mu^* = \sum_i p(M_i|D)\mu_i. \quad (7)$$

For example, one could choose the marginalization weights, $p(M_i|D)$, to be normally distributed according to the RMSE performance. We plan to investigate this approach in the near future.

Lastly, in Figure 3, we take a closer look at the timeseries results at site 2, the Blodgett Forest Station in CA. Here we can see an oscillatory artifact in the Noah land surface model. This is due to structural error in the model that does not account properly for snowmelt. Here we clearly see the ML model is able to correct the structural error and demonstrates the HPBM is able to learn and correct structural error and is not just correcting model biases.

## 6. CONCLUSIONS

Here we proposed a methodology for integrating ML and PBMs to create new dynamical model behaviors. To evaluate the performance, we created a HPBM to predict the top layer soil moisture states at a handful of AmeriFlux sites. At each site, we performed a one-year LOO validation and obtained as much as a 3-fold reduction in the average RMSE. These performance gains are a strong indication of the potential of this approach. We found some instability in performance believed to be a result of ignoring uncertainty propagation during training. We propose a simple weighted average procedure to reduce these effects, but ultimately plan to employ GP-SSMs.

In this work, we demonstrated the ability to learn model structural error using a HPBM at a particular geo-location. To build global models, we will need to create HPBM that can predict out-of-sample *both spatially and temporally*. In addition, we will need to include remote sensing data, and properly account for uncertainty propagation using GP-SSMs.